\pdfoutput=1
\documentclass[11pt]{article}

\usepackage[final]{lib/acl}     
\usepackage{graphicx}              
\usepackage{booktabs}              
\usepackage{times}
\usepackage{latexsym}
\usepackage[export]{adjustbox}
\usepackage[T1]{fontenc}
\usepackage[utf8]{inputenc}
\usepackage{microtype}
\usepackage{inconsolata}
\usepackage{graphicx}
\usepackage{longtable}
\usepackage{booktabs}  
\usepackage{enumitem}
\usepackage{amsmath}
\usepackage{tabularx}
\usepackage{subcaption}
\usepackage{listings}
\usepackage{tcolorbox}
\usepackage{enumitem}
\tcbuselibrary{breakable}
\usepackage{comment}
\usepackage{float}
\usepackage{dblfloatfix}
\usepackage{multirow}

\newtcolorbox{promptbox}[1][]{
  colback=blue!2,
  colframe=blue!50!black,
  boxrule=0.6pt,
  arc=3pt,
  left=6pt,
  right=6pt,
  top=6pt,
  bottom=6pt,
  fonttitle=\bfseries,
  coltitle=white,
  title=#1
}

\title{CHiRPE: A Step Towards Real-World Clinical NLP with Clinician-Oriented Model Explanations}

\author{
  Stephanie Fong\textsuperscript{1,2},
  Zimu Wang\textsuperscript{2,3},
  Guilherme C. Oliveira\textsuperscript{2},
  Xiangyu Zhao\textsuperscript{2},
  Yiwen Jiang\textsuperscript{2},
\\
  \textbf{Jiahe Liu\textsuperscript{2}},
  \textbf{Beau-Luke Colton\textsuperscript{1}},
  \textbf{Scott Woods\textsuperscript{4}},
  \textbf{Martha E. Shenton\textsuperscript{5}},
  \textbf{Barnaby Nelson\textsuperscript{1}},
\\
  \textbf{Zongyuan Ge\textsuperscript{2}},
  \textbf{Dominic Dwyer\textsuperscript{1,2,}\thanks{Corresponding author.}}
\\
  \textsuperscript{1}Orygen and The University of Melbourne \ \ \textsuperscript{2}AIM for Health Lab, Monash University \\
  \textsuperscript{3}University of Liverpool \ \ \textsuperscript{4}Yale School of Medicine, Yale University \\
  \textsuperscript{5}Brigham and Women's Hospital, Harvard Medical School
\\
     \texttt{stephanie.fong@unimelb.edu.au, dominic.dwyer@orygen.org.au}
}

\begin{document}
\maketitle

\begin{abstract}

The medical adoption of NLP tools requires interpretability by end users, yet traditional explainable AI (XAI) methods are misaligned with clinical reasoning and lack clinician input. We introduce CHiRPE (Clinical High-Risk Prediction with Explainability), an NLP pipeline that takes transcribed semi-structured clinical interviews to: (i) predict psychosis risk; and (ii) generate novel SHAP explanation formats co-developed with clinicians. Trained on 944 semi-structured interview transcripts across 24 international clinics of the AMP-SCZ study, the CHiRPE pipeline integrates symptom-domain mapping, LLM summarisation, and BERT classification. CHiRPE achieved over 90\% accuracy across three BERT variants and outperformed baseline models. Explanation formats were evaluated by 28 clinical experts who indicated a strong preference for our novel concept-guided explanations, especially hybrid graph-and-text summary formats. CHiRPE demonstrates that clinically-guided model development produces both accurate and interpretable results. Our next step is focused on real-world testing across our 24 international sites.   

\end{abstract}

\section{Introduction}
Mental illness is a major contributor to the global health burden \cite{mcgorry2025youth}, with psychotic disorders posing particular concern due to their relatively high mortality risk \cite{walker2015mortality}. Psychosis is often preceded by a prodromal Clinical High-Risk for Psychosis (CHR-P) phase \cite{wang2022identification}, during which early detection can enable interventions that may delay or prevent progression, improving long-term outcomes.

The PSYCHS (Positive SYmptoms and Diagnostic Criteria for the CAARMS-Harmonized SIPS) is a clinician-administered, semi-structured interview \cite{woods2024development} for detecting CHR-P. However, its routine use is constrained by substantial time and administrative effort \cite{oliver2022prognostic}. To address these limitations, recent advances in NLP offer a viable path forward. Automated analysis of interview transcripts can support scalable and consistent screening of disorder markers, thereby alleviating the reliance on labour-intensive manual assessment \cite{GARCIAMOLINA2024,na-etal-2025-survey}. However, their “black-box” nature significantly impedes adoption in clinical contexts where transparency and interpretability are essential \cite{topol2019high, cina2022we}.

\begin{figure*}[t]
    \centering
    \includegraphics[width=\textwidth]{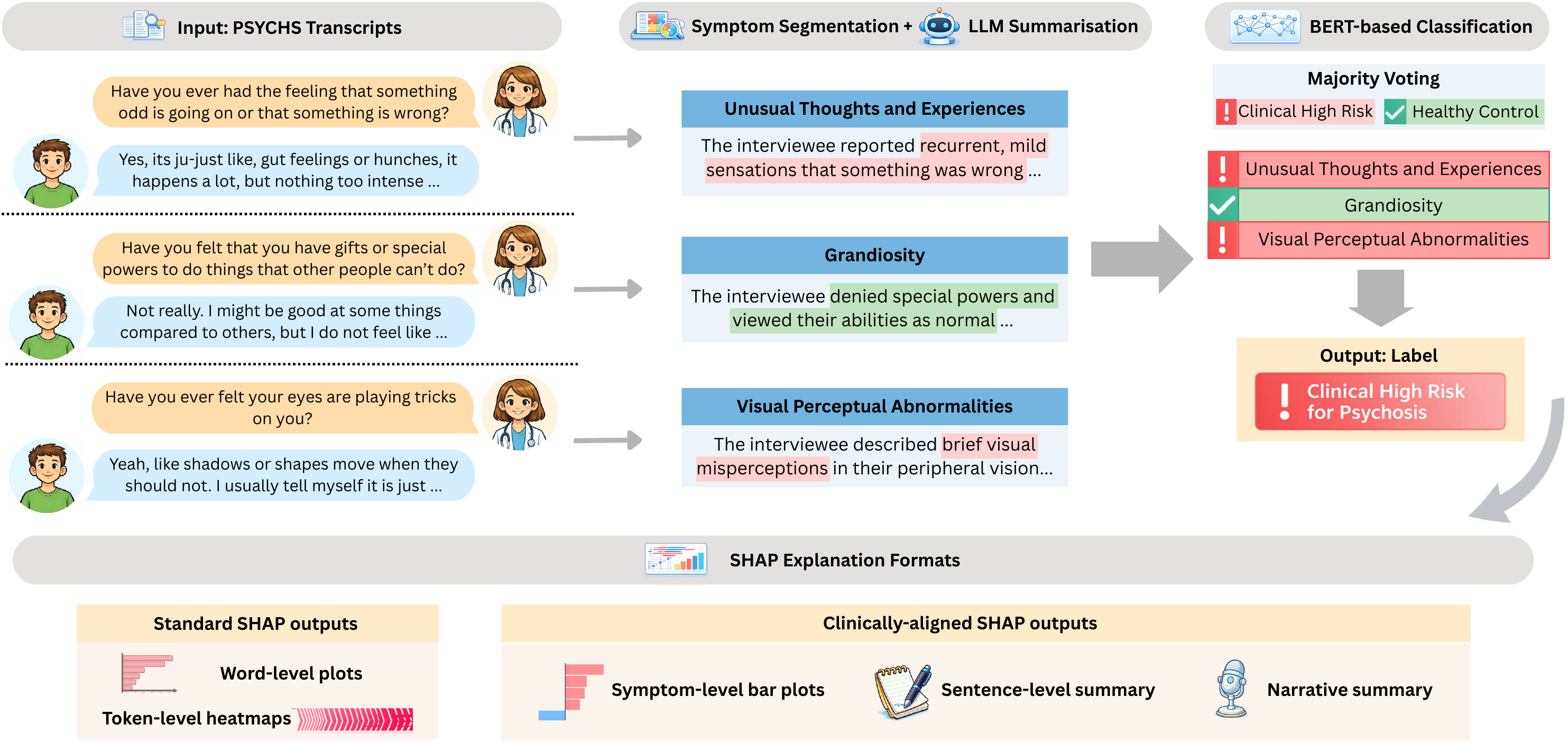}
    \caption{CHiRPE pipeline. Raw PSYCHS transcripts are segmented by symptom domain and summarised, then passed to a BERT-based classifier. The system outputs a CHR-P or Healthy label alongside SHAP explanations.}
    \label{fig:chirpe_pipeline}
\end{figure*}

Post-hoc explainable AI (XAI) techniques, such as SHapley Additive exPlanations (SHAP; \citealp{lundberg2017unified}), have been proposed to bridge this gap. In NLP, SHAP assigns each input feature (e.g. word or token) a value representing its average influence on the model's prediction \cite{shapley1953value}. While these attributions capture the model’s logic, they fundamentally misalign with real-world clinician reasoning \cite{lawrence2024opportunities}. In parallel, large language models (LLMs) remain opaque \cite{lawrence2024opportunities} and sensitive to subtle prompts or context changes \cite{peng2023does, zhou2023context, chatterjee2024posix}. Despite the importance of interpretability for clinical trust \cite{di2023explainable, talebi2024exploring}, clinician involvement in XAI design remains limited \cite{ghassemi2021false}.

Motivated by this gap, we introduce CHiRPE (Clinical High-Risk Prediction with Explainability), a human-centred NLP framework that processes transcribed PSYCHS interviews to classify individuals as CHR-P and generates five SHAP explanation formats, three of which are co-developed with clinicians (see Figure \ref{fig:chirpe_pipeline}). Trained on 943 transcripts across 24 international sites, the pipeline integrates symptom domain mapping, LLM summarisation, BERT-based classification, and SHAP-based explanations. Our key contributions are as follows: \textbf{(i)} an integrated pipeline CHR prediction that aligns model inputs with clinically meaningful constructs while preserving high classification accuracy; \textbf{(ii)} co-designed SHAP explanation formats grounded in how clinicians reason about psychosis risk; \textbf{(iii)} empirical clinician expert evaluation of these co-designed SHAP formats, showing that they outperform standard explanations in interpretability, clinical reasoning, and alignment.


\section{Related Work}
 A systematic review of nine studies using shallow classifiers with handcrafted features or static embeddings reported CHR-P diagnostic accuracies between 56-95\% \cite{GARCIAMOLINA2024}, but most relied on small, single-site samples and internal validation, raising concerns about overfitting and generalisability.
 Recent AMP-SCZ work using Llama 3 “normality” features and Na\"ive Bayes achieved moderate accuracy (AUC $\approx$ 0.68) on a smaller subset of the same PSYCHS interview data \cite{bilgrami2025leveraging}, yet lacked interpretability.

\begin{figure}[t]
  \centering
  \includegraphics[width=\linewidth]{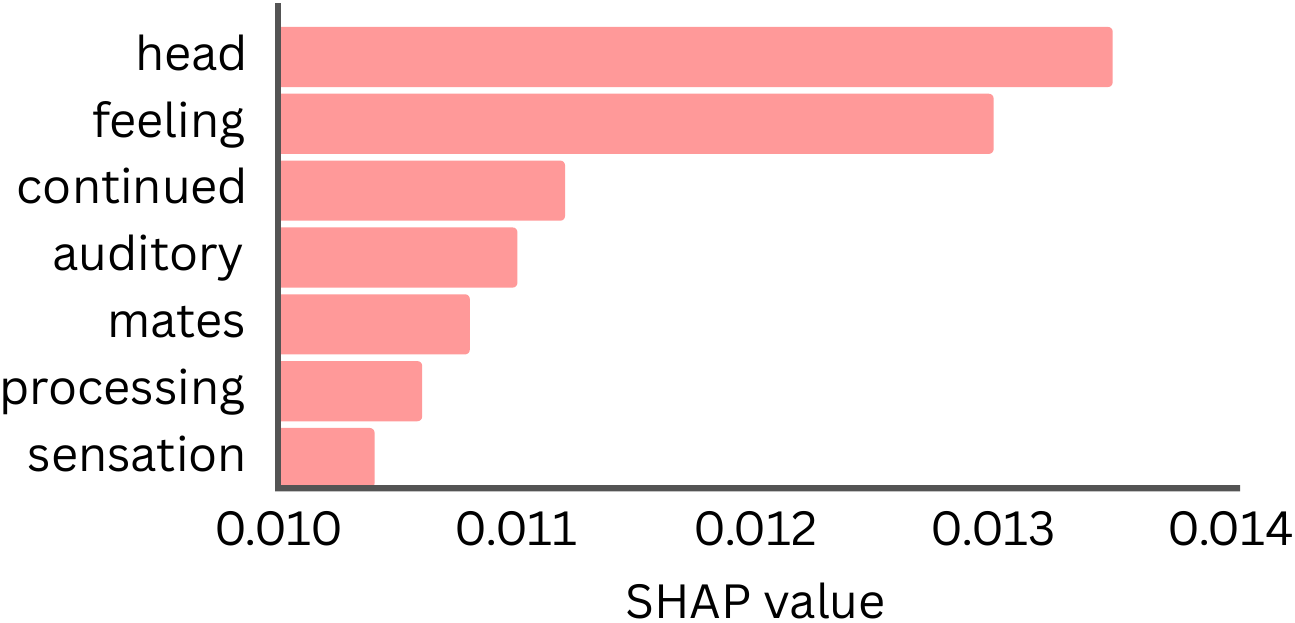}
  \caption{SHAP word-level bar plot showing the top contributing words for a CHR-P classification decision.}
  \label{fig:shap_token}
\end{figure}

Existing results demonstrate that the shift to deep learning enables richer linguistic modelling but introduces opacity, necessitating interpretability methods. While SHAP formats have shown promise in broader medical applications such as cerebral infarction \cite{nohara2022explanation} and thoracic surgery \cite{hur2025comparison}, mental health NLP studies remain limited \cite{baki2022multimodal, nie2025dual}. Standard SHAP word-level bar plots (Figure~\ref{fig:shap_token}) and token-level heatmaps (Figure~\ref{fig:shap_highlight}) either present decontextualised words in isolation or span entire transcripts, which are both misaligned with the types of concept-driven summaries used in clinical reasoning to make decisions. To our knowledge, no mental health NLP studies have examined adapted SHAP formats that align with clinical reasoning, and it remains a critical gap that is preventing the clinical use of NLP tools.

\section{CHiRPE}
The CHiRPE pipeline is presented in Figure \ref{fig:chirpe_pipeline}. It consists of the following: a) concept mapping to generate transcript segments; b) summarisation of concept segments; c) CHR detection classifier training; d) application of clinically-informed SHAP approaches. In this section, we introduce each step in detail.

\subsection{Symptom Domain Segmentation}
The PSYCHS interview follows a fixed order of standardised questions organised into 15 attenuated positive symptom domains
(see Appendix~\ref{app:template_questions}). Interviewer utterances were mapped to domains using fuzzy string matching\footnote{\url{https://pypi.org/project/fuzzywuzzy/}} with a threshold of 80\% (justified in Appendix \ref{app:fuzzy_segment}). 

\subsection{Summarisation of Interview Segments} 
Each interview segment was rephrased into third-person using a two-step process with Mistral-7B-Instruct-v0.3\footnote{\url{https://huggingface.co/mistralai/Mistral-7B-Instruct-v0.3}} to better match BERT pretraining data. The process involved an initial clinician style rewrite followed by refinement for completeness and coherence (see Appendix~\ref{app:summarise_prompt} for prompts and Appendix \ref{app:prompt_sensitivity} for prompt sensitivity analysis).

\subsection{Classification Models}
We fine-tuned BERT \cite{devlin2019bert}, ClinicalBERT \cite{alsentzer2019publicly}, and MentalBERT \cite{ji2021mentalbert}, domain-specific models for clinical and mental health language, to classify each summary symptom segment as CHR-P or control. Segments exceeding 512 tokens were split into fixed-length chunks, and transcript-level labels were determined by majority voting across segments.

\subsection{Explanation Generation with SHAP}
As baselines, the standard SHAP visualisations of word-level bar plots for top impact words and colour-coded heatmaps were generated, as shown in Figures~\ref{fig:shap_token} and ~\ref{fig:shap_highlight}. In consultation with clinicians, we introduced new presentation formats, including sentence-level summaries, narrative explanations, and symptom-level plots.

\paragraph{Sentence‐level Summary.}
For individuals classified as CHR, the sentence with the highest average net SHAP contribution toward a CHR prediction was extracted as a \emph{Sentence‐level Summary}. For example,
\begin{promptbox}[P4 Ideas of Guilt]

They admit to constantly dwelling on past problems, which significantly impact their daily life.

\end{promptbox}

\paragraph{Narrative Summary.}

A concise \emph{Narrative Summary} was generated using Qwen3-4B \cite{yang2025qwen3} for its strong contextual generation ability. Each summary included $\pm1$ sentence of context and a representative interviewee quote. The prompt is shown in Appendix \ref{app:narrative_summary}. An example summary is as follows:
\begin{promptbox}[P4 Ideas of Guilt]

 The interviewee reports persistent and intrusive thoughts centered on moral concerns and others’ perspectives, particularly during emotional distress. They describe frequent rumination on past issues, which interferes with daily functioning, and recurrent episodes of intense, unexplained guilt occurring at least weekly.

  \textit{``Even if it’s not actively thinking about the problem, like, ‘Oh, I wish I’d done something different,’ it affects little things in life that make you think about it.''}

\end{promptbox}

\begin{figure}[htbp]
  \centering
\includegraphics[width=\linewidth]{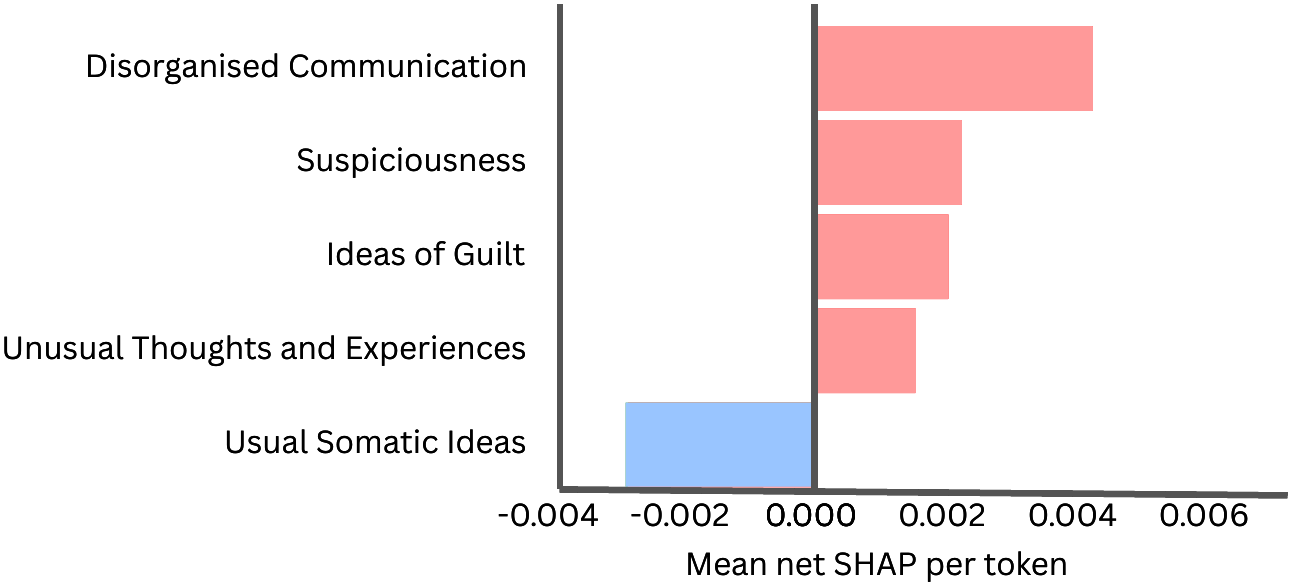}
  \caption{Symptom-level plots (red: CHR-P, blue: healthy control).}
  \label{fig:shap_symptom}
\end{figure}

\paragraph{Symptom-Level Plots.}
SHAP values were aggregated into mean net scores per symptom domain and visualised as horizontal bar charts (Figure \ref{fig:shap_symptom}), showing each domain's influence on CHR-P or control prediction.

\begin{table*}[htbp]
\centering
\small
\setlength{\tabcolsep}{3pt} 
\resizebox{\linewidth}{!}{\begin{tabular}{@{}lcccc|cccc|cccc|cccc@{}}
\toprule
\textbf{Model} 
& \multicolumn{4}{c}{\textbf{Baseline}} 
& \multicolumn{4}{c}{\textbf{Summary Only}} 
& \multicolumn{4}{c}{\textbf{Segmentation Only}} 
& \multicolumn{4}{c}{\textbf{Proposed (Summ + Seg)}} \\
\cmidrule(lr){2-5} \cmidrule(lr){6-9} \cmidrule(lr){10-13} \cmidrule(lr){14-17}
& Acc & F1 & Prec & Rec
& Acc & F1 & Prec & Rec
& Acc & F1 & Prec & Rec
& Acc & F1 & Prec & Rec \\
\midrule
BERT
& 83.91 & 89.78 & 83.11 & \textbf{97.62}
& 81.03 & 88.09 & 80.79 & 96.83
& \textbf{91.81} & \textbf{95.10} & 93.15 & 97.14
& 91.23 & 94.50 & \textbf{96.99} & 92.14 \\

ClinicalBERT
& 81.03 & 88.00 & 81.21 & \textbf{96.03}
& 82.76 & 88.97 & 82.88 & \textbf{96.03}
& 87.51 & 92.56 & 91.97 & 93.21
& \textbf{90.64} & \textbf{94.20} & \textbf{95.59} & 92.86 \\

MentalBERT
& 83.91 & 89.85 & 82.67 & 98.41
& 84.48 & 90.04 & 84.14 & 96.83
& \textbf{91.23} & 94.77 & \textbf{92.52} & 97.14
& \textbf{91.23} & \textbf{94.85} & 91.39 & \textbf{98.57} \\
\bottomrule
\end{tabular}}
\caption{Performance metrics of BERT-based models across ablation settings. Best results for each metric within each model are highlighted in bold.}
\label{tab:bert_results_horizontal}
\end{table*}
\section{Data Analysis Methods}

\subsection{Performance Metrics}
 Accuracy, precision, recall, F1, and AUC were evaluated at the transcript level on the test set. 

\subsection{Clinical Expert Feedback}
Clinical experts completed a mixed-method questionnaire on SHAP explanation preferences, including: a) word-level bar plots; b) colour-coded heatmaps; c) symptom-level bar plots; d) sentence-level summary; and e) narrative summaries (full questionnaire in Appendix~\ref{sec:questionnaire}). Descriptive and inferential statistics were used to compare interpretability ratings across formats.

\section{Data and Experimental Setup}
\subsection{Dataset}
Participants were drawn from the Accelerating Medicines Partnership Schizophrenia\footnote{\url{https://www.ampscz.org/}} (AMP-SCZ) study, focusing on CHR-P identification and transition to psychosis \cite{wannan2024accelerating}. The dataset comprised 943 English PSYCHS transcripts from 581 unique participants aged 12-30 (M = 20.9, SD = 4.1), 63.3\% of whom were female. Labels were assigned by trained research assistants, with 83.6\% CHR-P and 16.4\% Healthy Controls across 24 international sites of the AMP-SCZ study.

The dataset was partitioned into 64\% for training, 16\% for development (via nested 5-fold cross-validation on an 80\% subset), and 20\% for held-out testing. Splits were stratified by CHR status and grouped by participant ID to prevent data leakage. A fixed random seed ensured consistent splits across model variants.

\subsection{Pre-processing} Audio recordings were human transcribed\footnote{\url{https://www.transcribeme.com/}}, with all timestamps and personally identifiable information  removed. A few-shot XLM-RoBERTa classifier \cite{conneau2020unsupervised} distinguished interviewer and interviewee turns.

\subsection{Models and Hyperparameters}
BERT-based models were trained on a single A100 GPU within a
high-performance computing cluster.
All classification models were fine-tuned using weighted cross-entropy loss to address class imbalance. Hyperparameters, including learning rate, batch size, weight decay, and number of epochs, were optimised via inner-loop grid search (details in Appendix ~\ref{app:hyperparameter_table}).

\section{Results}

\subsection{Model Classification Performance}
All three transformer-based models performed strongly on the held-out test set (Table \ref{tab:bert_results_horizontal}; optimal hyperparameters in Appendix ~\ref{app:hyperparameter_table}), outperforming baselines without segmentation and summarisation. AUCs were 0.95 for BERT and 0.97 for both ClinicalBERT and MentalBERT, exceeding baseline AUCs of 0.94, 0.90, and 0.95, respectively. 

\subsection{Ablation Analysis}
\subsubsection{Segmentation and Summarisation}
Ablation results show that symptom domain segmentation is the main driver of performance gains, while summarisation provides smaller but consistent improvements (see Table \ref{tab:bert_results_horizontal}). In particular, segmentation reduces noise from long, heterogeneous transcripts and focuses the classifier on clinically meaningful information. By contrast, summarisation maintains comparable accuracy despite substantially reducing input length, improving modelling efficiency.

In the CHiRPE pipeline, segmentation and summarisation together achieve the best overall trade-off between F1, precision, and efficiency. The two components appear complementary: segmentation imposes clinically meaningful structure, while summarisation standardises and condenses content for SHAP attribution and downstream narrative generation, yielding both improved predictive performance and a more scalable explanation workflow.

\subsubsection{Precision–Recall Trade off}
As shown in Table \ref{tab:clinicalbert_recall_specificity}, while the proposed setting (summarisation + segmentation) improves F1 and precision, it yields a modest drop (-4\%) in recall and is accompanied by a substantial gain (+41\%) in specificity. From a modelling perspective, this pattern suggests reduced bias from class imbalance (CHR = 83.6\%, HC = 16.4\%) and less overfitting to CHR cases. Clinically, given that CHR reflects risk rather than diagnosis, and that false positives carry nontrivial clinical costs \cite{de2021probability, di2022antipsychotic}, this shift indicates a more conservative and better calibrated decision boundary.

\begin{table}[htbp]
\centering
\resizebox{\columnwidth}{!}{%
\begin{tabular}{lcc}
\toprule
\textbf{ } & \textbf{Recall (CHR)} & \textbf{Specificity (HC)} \\
\midrule
Baseline                   & 0.9603 & 0.3958 \\
Proposed (Summ + Seg)      & 0.9286 & 0.8065 \\
\bottomrule
\end{tabular}
}
\caption{Recall–specificity trade off for ClinicalBERT under ablation settings.}
\label{tab:clinicalbert_recall_specificity}
\end{table}

\subsection{Clinical Expert Evaluation of Explanation Formats}
Twenty-eight clinical experts, including psychiatrists, clinical psychologists, and trained research assistants,
completed a 10-minute questionnaire evaluating the interpretability of SHAP outputs on a 5-point Likert scale (Table \ref{tab:descriptive}). Responses were anonymous, and participants gave informed consent for data usage (Appendix ~\ref{sec:questionnaire}).

\begin{table}[htbp]
\centering
\resizebox{\columnwidth}{!}{%
\begin{tabular}{llc}
\toprule
\textbf{ } & \textbf{Explanation Format} & \textbf{Interpretability} \\
\midrule
\multirow{2}{*}{Baseline} 
 & Word-level plots     & 2.18 \\
 & Token-level heatmaps & 2.50 \\
\midrule
\multirow{5}{*}{Proposed} 
 & Symptom bar plots        & 3.71 \\
 & Sentence-level summary   & 4.25 \\
 & Single narrative         & 4.08 \\
 & Graph + single narrative & 4.32 \\
 & Multiple narratives      & 4.21 \\
\bottomrule
\end{tabular}
}
\caption{Mean interpretability ratings for baseline and proposed explanation formats (n = 28).}
\label{tab:descriptive}
\end{table}

\subsubsection{Quantitative Clinical Expert Feedback}
All novel SHAP formats outperformed the traditional SHAP word-level bar plots and heatmaps. The combination of symptom-level plots with narrative summaries received the highest interpretability ratings. A repeated-measures ANOVA confirmed significant differences across formats, $F(6, 28) = 26.485$, $p < .001$. Full ANOVA results are presented in Appendix~\ref{app:clinician_ratings}.

Additional ratings on Clinical Reasoning and Clinical Intuition Alignment showed the same pattern of results, with proposed formats consistently outperforming baseline visualisations. Full quantitative results are reported in Appendix~\ref{app:clinician_ratings}. 

\subsubsection{Qualitative Clinical Expert Feedback}
Qualitative coding of 20 clinical expert comments identified eight themes, most commonly favouring hybrid graph-text formats and concise presentations. Representative quotes are provided in Appendix \ref{app:clinician_ratings}.

\begin{table}[htbp]
\centering
\small
\begin{tabular}{lcc}
\toprule
\textbf{Theme} & \textbf{n} & \textbf{\%} \\
\midrule
Graph + text summaries preferred & 6 & 30 \\
Concise formats (dot points, fragments) & 4 & 20 \\
Structured symptom tables needed & 3 & 15 \\
Clear definition of experiences & 2 & 10 \\
Clarify model inputs and outputs & 2 & 10 \\
Overview across multiple symptoms & 1 & 5 \\
Add risk severity scale for decisions & 1 & 5 \\
Data access and governance issues & 1 & 5 \\
\bottomrule
\end{tabular}
\caption{Themes from qualitative feedback.}
\end{table}


\section{Conclusion and Future Work}
 In conclusion, CHiRPE outperformed baseline models in classification and surpassed existing SHAP approaches in interpretability. The preference for hybrid graph–text summaries and concise text formats highlighted CHiRPE’s practical contributions to future explainable clinical AI designs. CHiRPE addresses a key gap in mental health NLP in light of evolving AI regulations related to interpretability (e.g., the EU AI Act).
 
 While recent AMP-SCZ work using Llama 3 features achieved moderate accuracy (AUC $\approx$ 0.68) compared to CHiRPE's higher performance (AUC > 0.95) \cite{bilgrami2025leveraging}, differing methods and limited reporting in evaluation metrics preclude direct comparison. Prior work has also shown performance gains from domain-specific BERT models in mental health tasks \cite{ji2021mentalbert,turchin2023comparison}, but CHiRPE achieved comparable accuracy without task-specific pretraining. If replicated, this observation potentially suggests a need to shift focus from continual fine-tuning to aligning models with clinical reasoning, particularly in areas where diagnoses rely on subjective narratives rather than objective biomarkers \cite{oliver2022prognostic}.

 Future work will focus on refining CHiRPE's text-based summaries and symptom-level visualisations and piloting an interface prototype (Figure~\ref{fig:enter-label} in Appendix ~\ref{app:additional_fig}) across our network of 24 international sites. We will also incorporate patient perspectives to enhance concept mappings and extend the pipeline to additional mental health conditions, such as depression, mania, and anorexia.      

\section*{Limitations}
This study was limited to English transcripts collected from the 24 international sites. Future work could extend concept mapping and model evaluation  to multilingual data for broader applicability. Additional recruitment of participants in future studies could also strengthen the robustness and generalisability of findings. It is also worth exploring alternative models that are not BERT-based for this task. 

\section*{Ethical Considerations}
The transcript data used in this study were obtained from the Accelerating Medicines Partnership Schizophrenia (AMP-SCZ) consortium under a data use agreement. All data were de-identified and collected with informed consent for research purposes. Recruitment and consent procedures were IRB-approved \cite{ampscz2023protocol}. Participants will be reimbursed for their time and expenses associated with completing the research assessments. The amount of reimbursement varies by data collection site and the extent of study completion.

We used the pre-trained BERT model released by Google under the Apache License 2.0\footnote{\url{https://www.apache.org/licenses/LICENSE-2.0}} \cite{devlin2019bert}, ClinicalBERT under the MIT license\footnote{\url{https://tlo.mit.edu/understand-ip/exploring-mit-open-source-license-comprehensive-guide}} \cite{alsentzer2019publicly}, and MentalBERT under the Creative Commons Attribution Non Commercial 4.0 license\footnote{\url{https://creativecommons.org/licenses/by-nc/4.0/deed.en}} \cite{ji2021mentalbert}. All models were fine-tuned on AMP-SCZ data solely for research purposes, in accordance with their respective licenses.

The development of explainable AI tools for sensitive mental health settings raises important ethical and privacy challenges, particularly around informed consent. In clinical contexts where ambient data capture and AI-driven interpretation are involved, ethical safeguards cannot be assumed—they must be explicitly designed, tested, and adapted to the needs of each case. As part of our design process, we are co-developing CHiRPE’s consent procedures with both patients and clinicians to ensure transparency, agency, and appropriateness. This form of human-centred ethical design is expected to further enhance model performance, explanatory capacity, and ultimately the trust required for CHiRPE to contribute to the long-term sustainability of AI technologies in mental health care.

\section*{Data and Code Availability}
The AMP-SCZ dataset used in this study is available to researchers through the National Data Archive of the National Institute of Health in the USA: \url{https://www.ampscz.org/scientists/data/}. All code used for data preprocessing, model training, and explanation generation is available at \url{https://github.com/stephaniesyfong/CHiRPE}.

\section*{Use of AI}
ChatGPT and GitHub Copilot were used to assist with code debugging and language editing. All outputs were manually reviewed and verified by the authors.

\section*{Acknowledgements}
The Accelerating Medicines Partnership Schizophrenia (AMP-SCZ) is a public-private partnership managed by the Foundation for the National Institutes of Health. The AMP-SCZ research program is supported by contributions from the AMP-SCZ public and private partners, including NIMH (U24MH124629, U01MH124631, and U01MH124639) and Wellcome (220664/Z/20/Z and 220664/A/20/Z). The views expressed in this article are personal views of the authors and may not be understood or quoted as being made on behalf of or reflecting the position of the Department of Health and Human Services, including the National Institutes of Health and the US Food and Drug Administration, the United States Government, or the European Medicines Agency or one of its committees or working parties. The research was supported by the University of Melbourne's Research Computing Services and the Petascale Campus Initiative. The authors report no conflict of interest. Dwyer was supported by an NHMRC Emerging Leadership II Investigator Grant (\#2034943).

\bibliography{custom}

@article{de2021probability,
  title={Probability of transition to psychosis in individuals at clinical high risk: an updated meta-analysis},
  author={De Pablo, Gonzalo Salazar and Radua, Joaquim and Pereira, Joana and Bonoldi, Ilaria and Arienti, Vincenzo and Besana, Filippo and Soardo, Livia and Cabras, Anna and Fortea, Lydia and Catalan, Ana and others},
  journal={JAMA psychiatry},
  volume={78},
  number={9},
  pages={970--978},
  year={2021},
  publisher={American Medical Association}
}

@article{di2022antipsychotic,
  title={Antipsychotic treatment in people at clinical high risk for psychosis: a narrative review of suggestions for clinical practice},
  author={Di Lisi, Alessandro and Pupo, Simona and Menchetti, Marco and Pelizza, Lorenzo},
  journal={Journal of Clinical Psychopharmacology},
  pages={10--1097},
  year={2022},
  publisher={LWW}
}

@article{oliver2022prognostic,
  title={Prognostic accuracy and clinical utility of psychometric instruments for individuals at clinical high-risk of psychosis: a systematic review and meta-analysis},
  author={Oliver, Dominic and Arribas, Maite and Radua, Joaquim and Salazar de Pablo, Gonzalo and De Micheli, Andrea and Spada, Giulia and Mensi, Martina Maria and Kotlicka-Antczak, Magdalena and Borgatti, Renato and Solmi, Marco and others},
  journal={Molecular Psychiatry},
  volume={27},
  number={9},
  pages={3670--3678},
  year={2022},
  publisher={Nature Publishing Group UK London}
}

@article{woods2024development,
  title={Development of the PSYCHS: Positive SYmptoms and Diagnostic Criteria for the CAARMS Harmonized with the SIPS},
  author={Woods, Scott W and Parker, Sophie and Kerr, Melissa J and Walsh, Barbara C and Wijtenburg, S Andrea and Prunier, Nicholas and Nunez, Angela R and Buccilli, Kate and Mourgues-Codern, Catalina and Brummitt, Kali and others},
  journal={Early Intervention in Psychiatry},
  volume={18},
  number={4},
  pages={255--272},
  year={2024},
  publisher={Wiley Online Library}
}

@article{talebi2024exploring,
  title={Exploring the performance and explainability of fine-tuned BERT models for neuroradiology protocol assignment},
  author={Talebi, Salmonn and Tong, Elizabeth and Li, Anna and Yamin, Ghiam and Zaharchuk, Greg and Mofrad, Mohammad RK},
  journal={BMC Medical Informatics and Decision Making},
  volume={24},
  number={1},
  pages={40},
  year={2024},
  publisher={Springer}
}

@article{cina2022we,
  title={Why we do need explainable ai for healthcare},
  author={Cin{\`a}, Giovanni and R{\"o}ber, Tabea and Goedhart, Rob and Birbil, Ilker},
  journal={arXiv preprint arXiv:2206.15363},
  year={2022}
}

@article{di2023explainable,
  title={Explainable AI for clinical and remote health applications: a survey on tabular and time series data},
  author={Di Martino, Flavio and Delmastro, Franca},
  journal={Artificial Intelligence Review},
  volume={56},
  number={6},
  pages={5261--5315},
  year={2023},
  publisher={Springer}
}

@article{topol2019high,
  title={High-performance medicine: the convergence of human and artificial intelligence},
  author={Topol, Eric J},
  journal={Nature medicine},
  volume={25},
  number={1},
  pages={44--56},
  year={2019},
  publisher={Nature Publishing Group US New York}
}

@article{lundberg2017unified,
  title={A unified approach to interpreting model predictions},
  author={Lundberg, Scott M and Lee, Su-In},
  journal={Advances in neural information processing systems},
  volume={30},
  year={2017}
}

@article{ghassemi2021false,
  title={The false hope of current approaches to explainable artificial intelligence in health care},
  author={Ghassemi, Marzyeh and Oakden-Rayner, Luke and Beam, Andrew L},
  journal={The Lancet Digital Health},
  volume={3},
  number={11},
  pages={e745--e750},
  year={2021},
  publisher={Elsevier}
}

@article{lawrence2024opportunities,
  title={The opportunities and risks of large language models in mental health},
  author={Lawrence, Hannah R and Schneider, Renee A and Rubin, Susan B and Matari{\'c}, Maja J and McDuff, Daniel J and Bell, Megan Jones},
  journal={JMIR Mental Health},
  volume={11},
  number={1},
  pages={e59479},
  year={2024},
  publisher={JMIR Publications Inc., Toronto, Canada}
}

@article{ji2021mentalbert,
  title={Mentalbert: Publicly available pretrained language models for mental healthcare},
  author={Ji, Shaoxiong and Zhang, Tianlin and Ansari, Luna and Fu, Jie and Tiwari, Prayag and Cambria, Erik},
  journal={arXiv preprint arXiv:2110.15621},
  year={2021}
}

@article{turchin2023comparison,
  title={Comparison of BERT implementations for natural language processing of narrative medical documents},
  author={Turchin, Alexander and Masharsky, Stanislav and Zitnik, Marinka},
  journal={Informatics in Medicine Unlocked},
  volume={36},
  pages={101139},
  year={2023},
  publisher={Elsevier}
}

@article{chatterjee2024posix,
  title={POSIX: A Prompt Sensitivity Index For Large Language Models},
  author={Chatterjee, Anwoy and Renduchintala, HSVNS Kowndinya and Bhatia, Sumit and Chakraborty, Tanmoy},
  journal={arXiv preprint arXiv:2410.02185},
  year={2024}
}

@article{zhou2023context,
  title={Context-faithful prompting for large language models},
  author={Zhou, Wenxuan and Zhang, Sheng and Poon, Hoifung and Chen, Muhao},
  journal={arXiv preprint arXiv:2303.11315},
  year={2023}
}

@misc{ampscz2023protocol,
  title        = {Accelerating Medicines Partnership{\textregistered} Schizophrenia Observational Study Protocol},
  author       = {{Accelerating Medicines Partnership}},
  year         = {2023},
  howpublished = {\url{https://cdn.clinicaltrials.gov/large-docs/03/NCT05905003/Prot_000.pdf}},
  note         = {Accessed May 18, 2025},
  institution  = {ClinicalTrials.gov}
}

@article{wannan2024accelerating,
  title={Accelerating Medicines Partnership{\textregistered} Schizophrenia (AMP{\textregistered} SCZ): rationale and study design of the largest global prospective cohort study of clinical high risk for psychosis},
  author={Wannan, Cassandra MJ and Nelson, Barnaby and Addington, Jean and Allott, Kelly and Anticevic, Alan and Arango, Celso and Baker, Justin T and Bearden, Carrie E and Billah, Tashrif and Bouix, Sylvain and others},
  journal={Schizophrenia bulletin},
  volume={50},
  number={3},
  pages={496--512},
  year={2024},
  publisher={Oxford University Press US}
}

@article{alsentzer2019publicly,
  title={Publicly available clinical BERT embeddings},
  author={Alsentzer, Emily and Murphy, John R and Boag, Willie and Weng, Wei-Hung and Jin, Di and Naumann, Tristan and McDermott, Matthew},
  journal={arXiv preprint arXiv:1904.03323},
  year={2019}
}

@inproceedings{devlin2019bert,
  title={Bert: Pre-training of deep bidirectional transformers for language understanding},
  author={Devlin, Jacob and Chang, Ming-Wei and Lee, Kenton and Toutanova, Kristina},
  booktitle={Proceedings of the 2019 conference of the North American chapter of the association for computational linguistics: human language technologies, volume 1 (long and short papers)},
  pages={4171--4186},
  year={2019}
}

@article{shapley1953value,
  title={A value for n-person games},
  author={Shapley, Lloyd S and others},
  year={1953},
  publisher={Princeton University Press Princeton}
}

@article{nie2025dual,
  title={A Dual-Channel Prediction-Interpretation Framework with Pre-Trained Language Models and SHAP Explainability},
  author={Nie, Hui and Wu, Xiaoyan},
  journal={Journal of Computer and Communications},
  volume={13},
  number={3},
  pages={116--137},
  year={2025},
  publisher={Scientific Research Publishing}
}

@article{baki2022multimodal,
  title={A multimodal approach for mania level prediction in bipolar disorder},
  author={Baki, P{\i}nar and Kaya, Heysem and {\c{C}}ift{\c{c}}i, Elvan and G{\"u}le{\c{c}}, H{\"u}seyin and Salah, Albert Ali},
  journal={IEEE Transactions on Affective Computing},
  volume={13},
  number={4},
  pages={2119--2131},
  year={2022},
  publisher={IEEE}
}

@inproceedings{na-etal-2025-survey,
    title = "A Survey of Large Language Models in Psychotherapy: Current Landscape and Future Directions",
    author = "Na, Hongbin  and
      Hua, Yining  and
      Wang, Zimu  and
      Shen, Tao  and
      Yu, Beibei  and
      Wang, Lilin  and
      Wang, Wei  and
      Torous, John  and
      Chen, Ling",
    editor = "Che, Wanxiang  and
      Nabende, Joyce  and
      Shutova, Ekaterina  and
      Pilehvar, Mohammad Taher",
    booktitle = "Findings of the Association for Computational Linguistics: ACL 2025",
    month = jul,
    year = "2025",
    pages = "7362--7376",
    ISBN = "979-8-89176-256-5"
}

@article{GARCIAMOLINA2024,
  title={Automatic speech recognition in psychiatric interviews: a rocket to diagnostic support in psychosis},
  author={Molina, Jos{\'e} Tom{\'a}s Garc{\'\i}a and Gaspar, Pablo A and Figueroa-Barra, Alicia},
  journal={Revista Colombiana de Psiquiatr{\'\i}a (English ed.)},
  volume={54},
  number={4},
  pages={624--631},
  year={2025},
  publisher={Elsevier}
}

@article{wang2022identification,
  title={Identification and predictive analysis for participants at ultra-high risk of psychosis: A comparison of three psychometric diagnostic interviews},
  author={Wang, Peng and Yan, Chuan-Dong and Dong, Xiao-Jie and Geng, Lei and Xu, Chao and Nie, Yun and Zhang, Sheng},
  journal={World journal of clinical cases},
  volume={10},
  number={8},
  pages={2420},
  year={2022}
}

@article{hur2025comparison,
  title={Comparison of SHAP and clinician friendly explanations reveals effects on clinical decision behaviour},
  author={Hur, Sujeong and Lee, Yura and Park, Joongheum and Jeon, Yeong Jeong and Cho, Jong Ho and Cho, Duck and Lim, Dobin and Hwang, Wonil and Cha, Won Chul and Yoo, Junsang},
  journal={npj Digital Medicine},
  volume={8},
  number={1},
  pages={578},
  year={2025},
  publisher={Nature Publishing Group UK London}
}

@article{nohara2022explanation,
  title={Explanation of machine learning models using shapley additive explanation and application for real data in hospital},
  author={Nohara, Yasunobu and Matsumoto, Koutarou and Soejima, Hidehisa and Nakashima, Naoki},
  journal={Computer Methods and Programs in Biomedicine},
  volume={214},
  pages={106584},
  year={2022},
  publisher={Elsevier}
}

@article{mcgorry2025youth,
  title={The youth mental health crisis: analysis and solutions},
  author={McGorry, Patrick and Gunasiri, Hasini and Mei, Cristina and Rice, Simon and Gao, Caroline X},
  journal={Frontiers in Psychiatry},
  volume={15},
  pages={1517533},
  year={2025},
  publisher={Frontiers Media SA}
}

@article{walker2015mortality,
  title={Mortality in mental disorders and global disease burden implications: a systematic review and meta-analysis},
  author={Walker, Elizabeth Reisinger and McGee, Robin E and Druss, Benjamin G},
  journal={JAMA psychiatry},
  volume={72},
  number={4},
  pages={334--341},
  year={2015},
  publisher={American Medical Association}
}

@article{bilgrami2025leveraging,
  title={Leveraging AI and Language Analysis to Predict Psychosis Risk in Clinical High-Risk Individuals},
  author={Bilgrami, Zarina and Corcoran, Cheryl and Cecchi, Guillermo and Wolff, Phillip},
  journal={Biological Psychiatry},
  volume={97},
  number={9},
  pages={S68--S69},
  year={2025},
  publisher={Elsevier}
}

@article{peng2023does,
  title={When does in-context learning fall short and why? a study on specification-heavy tasks},
  author={Peng, Hao and Wang, Xiaozhi and Chen, Jianhui and Li, Weikai and Qi, Yunjia and Wang, Zimu and Wu, Zhili and Zeng, Kaisheng and Xu, Bin and Hou, Lei and others},
  journal={arXiv preprint arXiv:2311.08993},
  year={2023}
}

@article{yang2025qwen3,
  title={Qwen3 technical report},
  author={Yang, An and Li, Anfeng and Yang, Baosong and Zhang, Beichen and Hui, Binyuan and Zheng, Bo and Yu, Bowen and Gao, Chang and Huang, Chengen and Lv, Chenxu and others},
  journal={arXiv preprint arXiv:2505.09388},
  year={2025}
}

@inproceedings{conneau2020unsupervised,
  title={Unsupervised cross-lingual representation learning at scale},
  author={Conneau, Alexis and Khandelwal, Kartikay and Goyal, Naman and Chaudhary, Vishrav and Wenzek, Guillaume and Guzm{\'a}n, Francisco and Grave, Edouard and Ott, Myle and Zettlemoyer, Luke and Stoyanov, Veselin},
  booktitle={Proceedings of the 58th annual meeting of the association for computational linguistics},
  pages={8440--8451},
  year={2020}
}

\appendix


\clearpage
\onecolumn
\appendix
\section{PSYCHS template questions and symptom-domain matching}
\label{app:template_questions}
\begin{promptbox}[Sample list of P1–P3 standard template questions]
\small

  \textbf{P1 Unusual Thoughts and Experiences}
  \begin{itemize}[leftmargin=*, itemsep=1pt, topsep=2pt, parsep=1pt, partopsep=2pt]
  \item Have you ever had the feeling that something odd is going on or that something is wrong?
  \item Have you ever been confused at times whether something you have experienced is real or imaginary?
  \item Have you ever daydreamed a lot or found yourself preoccupied with stories, fantasies, or ideas?
  \item Has your experience of time ever seemed to have changed? Has it become unnaturally faster or unnaturally slower?
  \item Have you ever seemed to live through events exactly as you have experienced them before?
  \item Have familiar people or surroundings ever seemed strange?
  \item Have you felt that you or others or the world have changed in some way?
  \item Have you ever felt that you might not actually exist? Or that the world might not exist?
  \item Have you ever felt you can predict the future?
  \item Have you felt that things that were happening around you had a special meaning just for you?
  \item Have you ever felt the radio or TV or other electronic devices are communicating directly with you?
  \item Do you know what it means to be superstitious? Have you been superstitious?
  \item Have you ever felt that some person or force outside yourself has been controlling or interfering with your thoughts, feelings, actions or urges?
  \item Have you ever felt that ideas or thoughts that are not your own have been put into your head? Or that your own thoughts have been taken out of your head?
  \item Have your thoughts ever been broadcast so that other people know what you are thinking? Or ever said out loud so that other people can hear them?
  \item Have you ever thought that people might be able to read your mind? Or that you could read other people’s minds?
\end{itemize}

\textbf{P2 Suspiciousness}
\begin{itemize}[leftmargin=*, itemsep=1pt, topsep=2pt, parsep=1pt, partopsep=2pt]
  \item Have you ever felt like people have been talking about you, laughing at you or thinking about you in a negative way? 
  \item Have you ever found yourself feeling mistrustful or suspicious of other people? 
  \item Have you ever felt that you have to pay close attention to what's going on around you in order to feel safe? 
  \item Have you ever felt like you are being singled out or watched? 
  \item Has anybody been giving you a hard time or trying to hurt you? Do you have a sense of who that might be? Do you feel they have hostile or negative intentions?
\end{itemize}

\textbf{P3 Unusual Somatic Ideas}
  \begin{itemize}[leftmargin=*, itemsep=1pt, topsep=2pt, parsep=1pt, partopsep=2pt]
  \item Have you ever worried that something might be wrong with your body, your health, or a part of your body? Have you thought that it seems different to others in some way? 
  \item Have you worried about your body shape? 
  \item Have you ever worried that something odd is going on with your body that you can't explain?
\end{itemize}

\end{promptbox}

\begin{table*}[htp]
  \small
  \centering
  \label{tab:psychs_mapping}
  \resizebox{\textwidth}{!}{%
    \begin{tabular}{@{}l p{5cm} p{9cm}@{}}
      \toprule
      \textbf{Domain} & \textbf{Template Question} & \textbf{Matched Utterance (segment)} \\
      \midrule
      P1 & Have you experienced odd or unusual beliefs that other people find strange? 
         & Interviewer: I’m wondering if you’ve ever held beliefs that others might see as a bit unusual.  \\
      & 
         & Interviewee: [silence]  \\
      & 
         & Interviewer: For instance, do you ever feel like everyday events—say, a news headline—carry a secret message meant just for you?  \\
      &
         & Interviewee: Sometimes I feel like casual conversations on TV are speaking directly to me.  \\

      \addlinespace
      P2 & Have you ever heard or seen things that other people couldn’t perceive? 
         & Interviewer: Do you ever notice sounds or sights that others around you don’t seem to pick up on?  \\
      & 
         & Interviewee: [silence]  \\
      & 
         & Interviewer: Like hearing soft voices in an empty room, or glimpsing shadows that no one else sees?  \\
      &
         & Interviewee: Yes, sometimes I catch voices calling my name in an empty room.  \\

      \bottomrule
    \end{tabular}%
  }
  \caption{Symptom–question mapping example for PSYCHS domains P1 and P2 with interviewer segments.}
\end{table*}

\clearpage
\twocolumn

\twocolumn

\section{Segmentation threshold with Fuzzy Matching}
\label{app:fuzzy_segment}
A random subset of fifty transcripts was manually segmented into the 15 PSYCHS symptom domains by two human expert raters. The resulting gold-standard segments were compared with automatically generated ones using fuzzy matching at similarity thresholds of 70\%, 80\%, and 90\%. The 80\% threshold achieved best performance (see Table below) and was therefore used in our subsequent analyses.

\begin{table}[htbp]
\centering
\begin{tabular}{lccc}
\toprule
\textbf{Threshold} & \textbf{Precision} & \textbf{Recall} & \textbf{F1} \\
\midrule
70\% & 0.628 & 0.512 & 0.341 \\
80\% & 0.869 & 0.881 & 0.817 \\
90\% & 0.869 & 0.876 & 0.817 \\
\bottomrule
\end{tabular}
\caption{Averaged Macro Performance at different thresholds.}
\label{tab:threshold_perf}
\end{table}

\section{Prompt for Rephrasing and Summarisation of Transcripts}
\label{app:summarise_prompt}
\begin{promptbox}[First Pass (Initial Draft Prompt)]

You are an expert clinical interviewer. Summarise the following interview segment in a single third person paragraph, covering what was asked and the detailed response.  
\\
Interview segment: \textit{<segment>}  
\\
Draft summary:
\\

\end{promptbox}

\begin{promptbox}[Second Pass (Refinement Draft Prompt)]

Here is a transcript segment and an initial draft summary. Improve the summary by adding any important information from the segment that was missed. Keep third person narration, one coherent paragraph, and no bullet points.  
\\
Interview segment: \textit{<segment>}  
\\
Draft summary: \textit{<draft>}  
\\
Improved summary:

\end{promptbox}

\section{Hyperparameter tables} 
\label{app:hyperparameter_table}
We performed a grid search over the following ranges: learning rate {1e‑5, 2e‑5}, batch size {8, 16}, number of epochs {2, 3}, and weight decay {0.0, 0.01} for all models.

\begin{table}[H]
\centering
\resizebox{\columnwidth}{!}{
\begin{tabular}{lcccc}
\toprule
\textbf{Model} & \textbf{Learning Rate} & \textbf{Batch Size} & \textbf{Epochs} & \textbf{Weight Decay} \\
\midrule
BERT           & 2e-5         & 8           & 3       & 0.01          \\
ClinicalBERT   & 2e-5         & 8           & 3       & 0.01           \\
MentalBERT     & 2e-5         & 8           & 3       & 0.01          \\
\bottomrule
\end{tabular}
}
\caption{Best hyperparameters found via grid search.}
\label{tab:bert_hyperparameters}
\end{table}

\section{Prompt for Generating Narrative Summaries}
\label{app:narrative_summary}
\begin{promptbox}[Description based on Excerpt Summary:]

You are an expert clinical interviewer. Rewrite the excerpt into ONE clinician-friendly
paragraph (max 3 sentences) describing *symptom*.
\\
Excerpt: <segment>

\end{promptbox}

\begin{promptbox}[Description based on Excerpt Summary:]

Provide ONLY the interviewee quote (enclosed in double quotation marks) that clearly
illustrates *symptom* and supports "anchor". Output the quote and nothing else. \\
Transcript: <segment>\\
Quote:

\end{promptbox}

\section{Prompt Sensitivity for Transcript Summarisation}
\label{app:prompt_sensitivity}
Prompt sensitivity was evaluated with 3 different prompts, varying on role framing (experienced clinician, clinical assessor, clinical note writer), instruction specificity (ask simply to “summarise” versus explicitly specify including both questions and responses or improving clarity), and emphasis (conciseness, completeness, accuracy and structure). Summaries remained semantically consistent across prompts, with BERTScore F1 = 0.7558 and SentenceBERT cosine = 0.7758.

\section{SHAP-based stability}
We also evaluated the factual consistency and semantic quality of Qwen3-4B narrative summaries derived from symptom domains with highest SHAP attributions. Factual consistency was assessed using an NLI-based faithfulness check, showing 6.76\% contradicted statements. Semantic similarity remained high, with SentenceBERT cosine = 0.7100 and BERTScore F1 = 0.8883 (precision = 0.8754, recall = 0.9019).

\clearpage
\onecolumn
\section{Questionnaire for Explanation Feedback}
\label{sec:questionnaire}
\begin{promptbox}[Background Information and Data Collection Purpose]
\small

\textbf{About Chirpe}\\
In Orygen, our team produces Artificial Intelligence (AI) devices that enhance human connection and identify consumers who need specialised care. We have developed an AI called "Chirpe" (Clinical High Risk Prediction with Explainability). Chirpe is able to identify young people at clinical high-risk for psychosis (CHR) using recorded speech of the PSYCHS questionnaire at accuracies above 90\%. It's very exciting because it means that we may not need to rate the PSYCHs in the future.\\
\textbf{Why your feedback matters}\\
Chirpe sometimes struggles with explaining why the person is CHR and that's where we need your help. We have designed different ways that Chirpe can explain their decisions and we'd like you to let us know what style of communication you prefer. We'd be really grateful for any other feedback too as you are the experts. \\
\textbf{Contribute to research}\\
If you want to be involved in this paper as a clinical advisor, please let us know. We will also use your responses to design better ways that CHiRPE can talk with you.
\\
\end{promptbox}

\begin{figure}[!htbp]
  \centering
  \includegraphics[width=\textwidth]{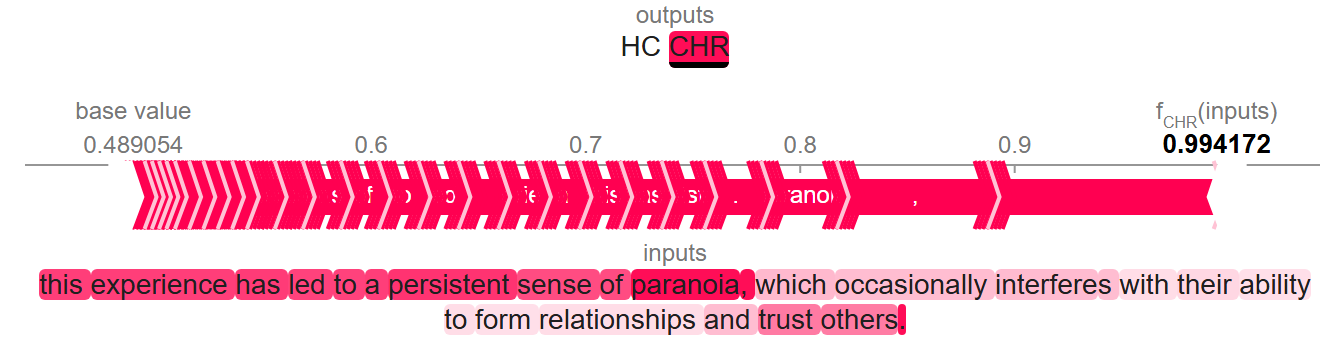}
  \caption{Inline token level heatmaps with SHAP}
  \label{fig:shap_highlight}
\end{figure}

\begin{promptbox}[Description of task]
\small

\textbf{Comprehensibility of Output Formats}\\
We are interested in how we can explain Chirpe's predictions to clinicians and have created 7 different options to choose from. Think about whether you understand why Chirpe made the decision: Is it comprehensible?\\

\textbf{Rating of Explanability Outputs}\\
Clinical experts were asked to evaluate the interpretability of the following seven explanation formats. For each, they rated interpretability on a 5-point Likert scale (1 = Not Interpretable; 5 = Very Interpretable).

\begin{enumerate}
    \item \textbf{Word-level SHAP plots} \\
    Longer bars in red indicate stronger influence towards CHR prediction (see Figure~\ref{fig:shap_token})
    
    \item \textbf{Highlighting text in the transcript or summary} \\
    Word highlights show influence (darker = stronger): red for CHR, blue for Control (see Figure~\ref{fig:shap_highlight})

    \item \textbf{Symptom-level Plots (P-items)} \\
    Red bars = CHR; Green bars = Controls.
    (See Figure~\ref{fig:shap_symptom} in Main Text)

    \item \textbf{Selected sentences from the most influential symptom categories} \\
    These sentences contain words that are most contributing to the CHR prediction. (See \textbf{Sentence-level Summary} in Main Text)

    \item \textbf{Single clinical narrative}
    (See \textbf{Narrative Summary} in Main Text)

    \item \textbf{Symptom-level Plots + Single clinical narrative}
    (See \textbf{Narrative Summary} in Main Text)

    \item \textbf{Multiple narrative summaries across symptom categories}
\end{enumerate}

\textbf{Qualitative Feedback}
\noindent In addition to the above ratings, clinicians responded to the following open-ended questions:

\begin{itemize}[itemsep=0pt, topsep=0pt, parsep=0pt]
    \item Would you suggest any changes to how the graphs are presented?
    \item Would you suggest any changes to how text information is presented?
    \item Do you have any other comments on how CHiRPE’s explanations can be improved?
\end{itemize}

\end{promptbox}
\clearpage
\twocolumn

\section{Clinician Ratings of Explanation Formats}
\label{app:clinician_ratings}

\subsection{ANOVA Results}
A one-way repeated-measures ANOVA revealed a significant effect of explanation format on perceived comprehensibility, $F(6, 28) = 26.485$, $p < .001$. Holm-adjusted pairwise $t$-tests showed that Options 4, 5, 6, and 7 were all rated significantly higher than Options 1 and 2 ($p < .001$). No significant differences were found between the top-rated formats (Options 4--7).

\subsection{Additional Evaluation with Clinical Reasoning and Alignment}
Two additional questions regarding clinical reasoning and alignment of SHAP explanation formats were evaluated using a 7-point Likert Scale.
\begin{enumerate}
    \item Clinical Reasoning: “To what extent does this {explanation format} align with how you would understand symptom information?”
    \item Clinical Intuition Alignment: “Does the {explanation format} present information in a way that is clinically intuitive?”
\end{enumerate}

\begin{table}[htbp]
\centering
\resizebox{\columnwidth}{!}{%
\begin{tabular}{lcc}
\toprule
\textbf{Explanation Format} & \textbf{Reasoning} & \textbf{Alignment} \\
\midrule
Word-level plots              & 2.0 & 1.8 \\
Token-level Heatmaps          & 2.4 & 2.4 \\
Symptom bar plots             & 4.2 & 4.4 \\
Sentence-level summary        & 5.4 & 5.6 \\
Single narrative              & 5.8 & 5.6 \\
Graph + Single narrative      & 6.0 & 6.2 \\
Multiple narratives           & 6.0 & 5.6 \\
\bottomrule
\end{tabular}
}
\caption{Mean and SD of Comprehensibility (n = 5)}
\label{tab:additional_descriptive}
\end{table}

Across these new evaluations, the proposed formats from this paper remained the highest rated, consistent with our earlier findings.

\subsection{Clinical Expert Quotes from Qualitative Feedback}

\begin{itemize} [itemsep=0pt, topsep=0pt, parsep=0pt]
\item\textbf{Graph plus text summaries preferred}\\
Clinical experts consistently valued mixed formats. One comment noted that “a graph combined with select sentences would be most useful” and another stated “I think a mixture of option six and seven would be perfect.” \\

\item\textbf{Concise formats} \\
Several clinical experts preferred brief explanations, e.g. “Can be more concise. No need to be complete sentences” and “Instead of a paragraph of text, I’d like dot points.” \\

\item\textbf{Preference for structured symptoms}\\
An expert emphasised alignment with clinical structure: “I prefer a summary and a table of all Ps with symptoms described through tenacity, distress, interference and frequency.” \\

\item\textbf{Clear definitions of experiences} \\
Two comments highlighted the need to define the target experience: “More specific description of the experience itself” and “still providing a global picture of several symptoms rather than only a single narrative.” \\

\item\textbf{Clarify model inputs and outputs}\\
Request of clarification of the model’s input, “I’m unclear if the system was given a full transcript or only the case summary.” \\

\item\textbf{Severity or risk indication} \\
Suggestion of adding a risk scale: “Could a scale of risk severity be included as a conclusion in the future.” \\

\item\textbf{Data access and governance}\\
One comment raised governance concerns: “Is the model in-house only. Who has access to these data.” 
\end{itemize}

\clearpage
\onecolumn
\section{Additional Figures and Tables}
\label{app:additional_fig}

\begin{figure}[!htbp]
    \centering
    \includegraphics[scale=0.4,frame]{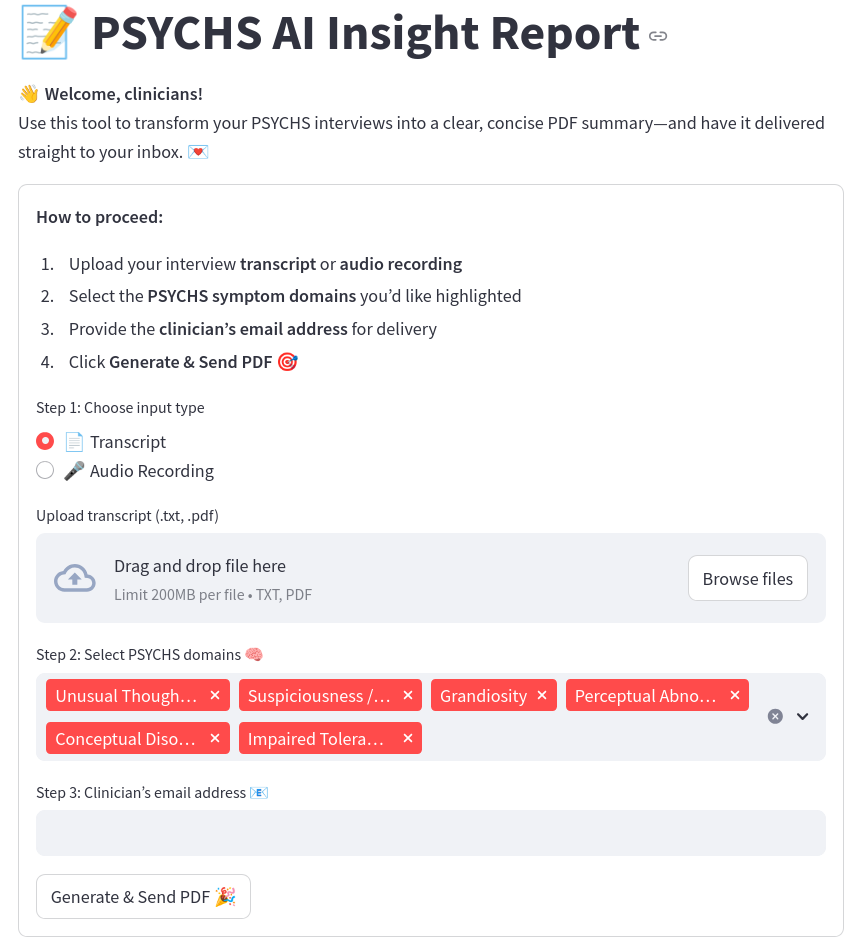}
    \caption{Website Interface.}
    \label{fig:enter-label}
\end{figure}

\end{document}